\documentclass{article}

\usepackage[nonatbib,preprint]{neurips_2020}

\usepackage[utf8]{inputenc} 
\usepackage[T1]{fontenc}    
\usepackage{hyperref}       
\usepackage{url}            
\usepackage{booktabs}       
\usepackage{amsfonts}       
\usepackage{nicefrac}       
\usepackage{microtype}      

\usepackage{algorithmic}
\usepackage{algorithm}

\usepackage{graphicx}
\graphicspath{ {./images/} }

\usepackage{biblatex}
\usepackage{amsmath}

\usepackage[table,xcdraw]{xcolor}

\usepackage{xcolor}

\title{EnK: Encoding time-information in convolution}

\author{%
  Avinash K Singh and Chin-Teng Lin\\
  Center for Artificial Intelligence, School of Computer Science\\
  University of Technology Sydney\\
    Australia \\
  \texttt{avinash.singh@uts.edu.au}\\
}

\begin{document}

\maketitle

\begin{abstract}
Recent development in deep learning techniques has attracted attention in decoding and classification in EEG signals. Despite several efforts utilizing different features of EEG signals, a significant research challenge is to use time-dependent features in combination with local and global features. There have been several efforts to remodel the deep learning convolution neural networks (CNNs) to capture time-dependency information by incorporating hand-crafted features, slicing the input data in a smaller time-windows, and recurrent convolution. However, these approaches partially solve the problem, but simultaneously hinder the CNN's capability to learn from unknown information that might be present in the data. To solve this, we have proposed a novel time encoding kernel (EnK) approach, which introduces the increasing time information during convolution operation in CNN. The encoded information by EnK lets CNN learn time-dependent features in-addition to local and global features. We performed extensive experiments on several EEG datasets: cognitive conflict (CC), physical-human robot collaboration (pHRC), P300 visual-evoked potentials, movement-related cortical potentials (MRCP). EnK outperforms the state-of-art by 12\% (F1 score). Moreover, the EnK approach required only one additional parameter to learn and can be applied to a virtually any CNN architectures with minimal efforts. These results support our methodology and show high potential to improve CNN performance in the context of time-series data in general. The sourcecode for EnK is available at https://github.com/thinknew/enk.

\end{abstract}

\section{Introduction}
\begin{figure}
\centering
\includegraphics[width=0.7\textwidth]{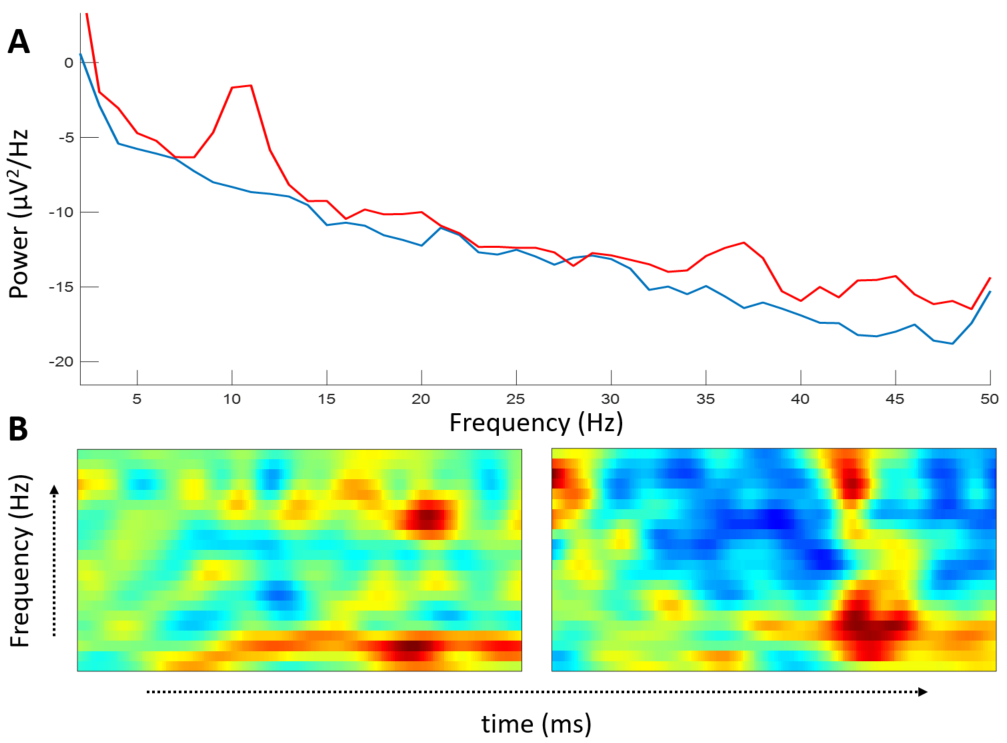}
\caption{ An example of time-dependent features. A) Power spectral density (PSD) of two EEG signals; B) event-related spectral perturbation (ERSP) of two EEG signals used in PSD }
\label{Figure 1}
\end{figure}

Electroencephalogram (EEG) is widely used in research involving neural engineering, cognitive neuroscience, neurotechnology, and brain-computer interface (BCI). EEG signals provide high temporal resolution, are non-invasive, and are relatively cheaper to run than other brain imaging techniques. In a typical scenario, the user is required to process EEG-signals to remove artifacts (eye, muscle, electrical noises, broken sensors), extract features (time-frequency domain, spectrograms, power ratios), and classify. No doubt, such work requires extensive domain knowledge and labor, on the top of work needed to record EEG signals. Therefore automating the whole process is essential particular with respect to BCI applications. Recent development in deep learning techniques has attracted attention among EEG researchers and race to develop a robust BCI. Despite several efforts utilizing different features of EEG signals in an automatic fashion, a significant research challenge is to use raw EEG data. The raw EEG data naturally come with time-dependent feature and such features are highly crucial for decoding and classifying EEG signals. Moreover, learning directly from time-dependent feature indirectly overcome the manual signal processing and feature extraction tasks. An example of time-dependent features in EEG signal is time-frequency information. Frequency information of a EEG signal alone can be seen as a feature. As shown in Figure \ref{Figure 1}(A), the frequency of two signals is peaking around 10 Hz, followed by similar but with smaller peaks around 20 Hz and 35Hz. What if combined the frequency with time? We converted two signals from Figure \ref{Figure 1}(A) into frequency over time, also popularly known as an event-related spectral perturbation (ERSP) \cite{makeig1993auditory} in EEG community. The transformed information is full of feature space, which can not reflected by two signals alone in Figure \ref{Figure 1}(A).This example clearly shows the importance of time-dependent features in EEG signals.

There has been a lot of research in EEG \cite{bashivan2015learning} \cite{li2020epileptic} \cite{zhang2019convolutional} in combining the time-dependency feature while learning local and global features. Despite that, the approaches are useful but often come with the cost of dedicated architecture to a specific task, are computationally expensive, and lack the ability to generalize to different tasks. To solve these problems, we introduce EnK, a kernel-based approach for convolution operation. EnK adds time-information into the EEG-signals while performing the convolution operation. This new information creates a feature space for time-dependent information that is generalized to any architecture and EEG-tasks. We have evaluated the efficacy and generability of EnK with different variety of EEG datasets collected from different tasks: cognitive conflict (CC) \cite{singh2018visual}, physical human-robot collaboration (pHRC) \cite{singh2020prediction}, P300 \cite{luck2014introduction}, and movement-related cortical potential (MRCP) \cite{shibasaki1980components}. These datasets have been collected from different settings and environments and vary in quality of signals, number of EEG-channels, size of datasets, and the number of participants. The main contributions of EnK are as follows:

\begin{itemize}
  \item EnK is a novel approach to encode the time-information in the data during the horizontal shifting of a kernel in convolution operation.
  \item EnK does not require any domain-specific knowledge or hand-crafted features, therefore automates the time-dependent feature extraction process.
  \item EnK is task-independent and architecture-independent; therefore, it can be applied to any new and existing CNNs architectures.
  \item The EnK approach requires one additional feature to learn and therefore is computationally inexpensive for an existing architecture.
\end{itemize}

\subsection{Related work}
The past few years have seen an increased number of deep learning applications in understanding and classifying EEG signals \cite{craik2019deep}. Deep learning has already shown a high number of successful applications in the field of natural language processing and computer vision, text classification, and action recognition \cite{lecun2015deep} \cite{deng2018deep} (review: \cite{canziani2016analysis}). A property of deep learning is to learn valuable information from raw data without manual labor \cite{chai2017improving}, which is very useful in the case of EEG signals. Convolutional neural networks (CNNs) are one of a most popular method in the field of deep learning and has proven effectiveness in several EEG based applications such as epilepsy/seizures prediction \cite{lu2019residual} \cite{emami2019seizure}, for detection of visual-evoked responses \cite{cecotti2010convolutional}, motor imagery classification  \cite{tayeb2019validating}, and speller \cite{shan2018simple}. Although deep learning can learn from raw EEG data, preprocessing of EEG signals is still required to reach optimal performance. These preprocessing methods are highly dependent on individual data sets and expert knowledge. Other than that, also vary by choice of filtering \cite{lotte2015signal}, channel referencing \cite{hu2018reference}, and artifact removal \cite{uriguen2015eeg} methods. Further, such preprocessed EEG signals also hinder deep learning models' ability to learn other relevant features, which might present in the data. A model that can learn from raw EEG data without hand-crafted preprocessing and feature extraction is highly desirable, particularly for BCI technologies. \par

There have been efforts towards learning time-dependency information using CNN. Pinheiro et al. \cite{pinheiro2014recurrent} \cite{liang2015recurrent} utilized the recurrent CNN (RCNN) to learn dependencies in context and learn the context in neighboring information. Inspired by RCNN, Bashivan et al. \cite{bashivan2015learning} trained EEG signals of mental workload using RCNN. The author showed that RCNN could learn spatial, spectral, and temporal features from EEG signals and improve the classification performance.  \par

Cui et al. \cite{cui2016multi} proposed a multi-scale CNN (MCNN) model for time-series data. MCNN automatically extracts features from identity mapping, down-sampling, and spectrogram and locally convolve them. The convolved output is then followed by concatenation into a full convolution to predict time-series data. Lea et al. \cite{lea2016temporal} presented a temporal convolution model (TCN) which learns video-based action first by learning an individual frame of video using CNN followed by a recurrent neural network (RNN). Although MCNN and TCN show a promising approach to capturing time-dependency features, it does not promise similar for EEG signals. \par

Following the trend of use of the recurrent and convolution model, Zhang et al. \cite{zhang2019convolutional} proposed a convolution recurrent attention model (CRAM) for EEG signal analysis. CRAM utilizes CNN to learn high-level representation in EEG signals, while recurrent attention mechanisms are used to learn temporal features in EEG signals. The model showed a significant improvement in classifying the motor-imagery based EEG signals. Recently, Li et al \cite{li2020epileptic} proposed another approach based on a unified temporal-spectral utilizing a group convolution squeeze-and-excitation network to detect epileptic seizures in EEG signals. This model claims to learn both spectral and temporal information from the epileptic seizure based EEG signals but again was dedicated to one task only. \par 

Inspired by several CNN models, the author \cite{schirrmeister2017deep} proposed a very deep convolution neural network model called DeepConvNet inspired from VGG \cite{simonyan2014very} for EEG signals. DeepConvNet demonstrated that it could learn different kinds of information in decoding task-related to EEG signals. The model shows promising results, but it highly depends on the size of input data to sufficiently converge and learn features.  \par 

Although several models have been proposed to learn time-dependency features primarily by combining recurrent and convolution networks, they are always dependent on the specific type of task to collect EEG signals. On the contrary, the use of convolution network alone without any recurrent model has shown  better success with decoding and classifying a variety of tasks in EEG signals. Recently, there was a more generalized model demonstrated by Lawhern et al. \cite{lawhern2018eegnet}, called EEGNet. The EEGNet is focused on a compact CNN model utilizing the depthwise and separable CNN \cite{chollet2017xception} approaches. EEGNet model has shown optimal performance from a variety of paradigms with EEG signals. The author also showed that EEGNet is very effective to generalize on different EEG signals. Although EEGNet is generalized for a variety of EEG tasks, it did not use any specific deep learning module to learn temporal features. It is highly reliant on the CNN approach, which is well known for learning local and global features only. 

\section{Materials and Methods}

\subsection{Data description}
We have used four different EEG datasets to evaluate our approach. The description of datasets is as below (also see Table \ref{Table 1}).

\subsubsection{CC and pHRC}
The cognitive conflict (CC) is an ERP elicited due to unexpected visual stimuli in EEG data. The visual stimuli are repeatedly presented to participants and asked to perform a certain tasks, and then a sudden change in expected behavior happens. Due to this, a negative deflection occurs 150-250ms in the frontal region of the brain known as PEN. There are two different kinds of EEG data used from a different kind of cognitive conflict task. In the first task, participants sit on the chair and perform the task in a controlled environment \cite{singh2018visual}, while in the second, a participant performs the task in a real-world environment with ANBOT \cite{aldini2019effect}\cite{singh2020prediction}. The goal is to classify conflict with non-conflict conditions.

\subsubsection{P300}
The P300 is an ERP elicited due to visual stimuli in EEG data. The visual stimuli are based on an oddball visual paradigm. In this paradigm, participants were shown a non-frequent "target" with frequent "non-target." The P300 waveform is a large positive deflection around 250-350ms on the parietal cortex whenever the target appears. The EEG data used here has been taken from BCI Competition III (Dataset II)\cite{blankertz2005bci}. The goal here is to classify EEG signals into the target with non-targets.

\subsubsection{MRCP}
Some neural activities contain both ERP as well as an oscillatory component. One particular example of this is the MRCP, which can be elicited by voluntary movements of the hands and feet. It is observable through EEG signals along the central and midline regions, contralateral to the hand or foot movement. The MRCP has been used previously to develop motor control BCIs for both healthy and physically disabled patients. The MRCP data used here is taken from BCI Competition II (Dataset IV)\cite{blankertz2004bci}. The goal here is to classify the four voluntary movements from hand and feet.

\begin{table}[h]
\centering
\caption{Data description}
\label{Table 1}
\begin{tabular}{lcccc}
\toprule
Datasets Name & Channels & Sampling rate & No. of classes & Dimension \\
\cmidrule{1-1} \cmidrule{2-2} \cmidrule{3-3} \cmidrule{4-4} \cmidrule{5-5} 
CC & 62 & 1000 & 2 & 62x1200x6841 \\
pHRC & 32 & 1000 & 2 & 32x1200x4895 \\
P300 & 64 & 240 & 2 & 64x240x340 \\
MRCP & 28 & 1000 & 4 & 28x500x316 \\
\bottomrule
\end{tabular}
\end{table}

\subsection{Classification method and evaluation}

\subsubsection{EnK: time-information encoding in CNN}

Convolution can be conceived as sliding the kernel along the input signal and taking a dot product with the corresponding portion at each location. A dot product in convolution can be represented by equation( \ref{equ1}). 

\begin{equation} \label{equ1}
 y[x,k]= \sum_{j=1}^{m} \sum_{i=1}^{n} x[i,j]*k[i,j]
\end{equation}

where signal is \(x\) and kernel \(k\) with dimension \(m\) and \(n\). Given dot operation \(y[x,k]\), a convolution operation on input data can be represented by equation (\ref{equ2}):

\begin{equation} \label{equ2}
Y
=
\begin{bmatrix}
y[x_{0,0},k+1*b] &  y[x_{0,1},k+2*b] & .......  & y[x_{0,n-1},k+n*b]\\
y[x_{1,0},k+1*b] &  y[x_{1,1},k+2*b] &....... & y[x_{1,n-1},k+n*b]\\
 ..  &   ..  &   ..  & ..  & \\
 ..  &   ..  &   ..  & ..  & \\
 y[x_{m-1,0},k+1*b] &  y[x_{m-1,1},k+2*b] &....... & y[x_{m-1,n-1},k+n*b]
\end{bmatrix}
\end{equation}

where $ 
    b= 
\begin{cases}
    0,& \text{for convolution}\\
    \in \mathbb{R}_{\neq 0},& \text{for EnK}
\end{cases}
$

\(b \neq 0\) represents the EnK which adds the time-information with scale \(b\) to the the kernel \(k\) on sliding over the column. \\

\begin{algorithm}
\caption{Calculate $EnK (Input, Kernel, h,w,kh,kw)$}
\begin{algorithmic}
\STATE $hout = h - kh +1$
\STATE $wout = w - kw +1$
\STATE $lincount = 0$
\FOR{$i \leq hout$}
\STATE $lincount = lincount+1$
\FOR{$j \leq wout$} 
\STATE $EnKernel = Kernel + lincount$ 
\STATE $elmul = Input[i:i+hout, j:j+wout] * EnKernel$
\STATE $Output(hout, wout) = sum(elmul)$
\ENDFOR
\ENDFOR
\end{algorithmic}
\end{algorithm}

\begin{figure}[h]
\centering
\includegraphics[width=1\textwidth]{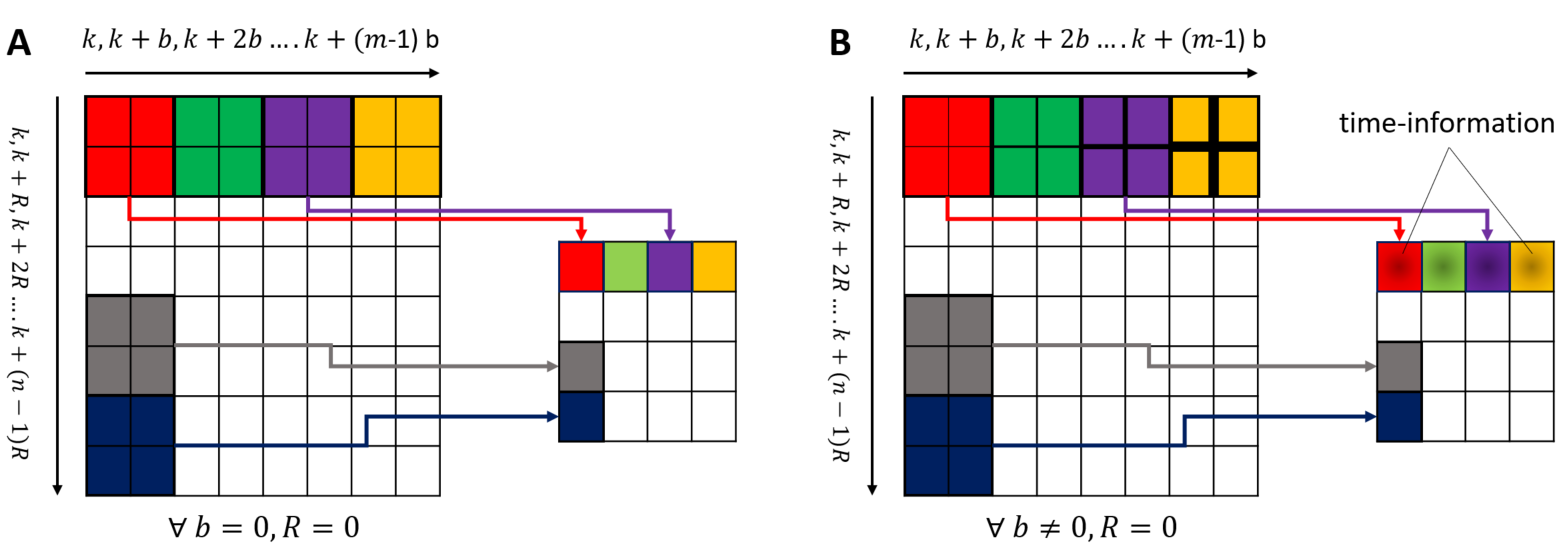}
\caption{Convolution. A. Standard convolution procedure where kernel value always stays the same when sliding over rows and column; B. EnK convolution operation where kernel value stays the same when sliding over rows but linearly increased when sliding over columns. ($k = kernel, m= number of rows for input, n = number of columns for input, C and R are scaling multipliers$)}
\label{Figure 2}
\end{figure}

Figure \ref{Figure 2} shows the standard convolution operation (left) and convolution operation with EnK (right). The standard convolution follows regular convolution operation while convolution operation with EnK adds time-information over the column (also see Algorithm 1).

\subsection{Baseline methods}
\begin{figure}[h]
\centering
\includegraphics[width=0.8\textwidth]{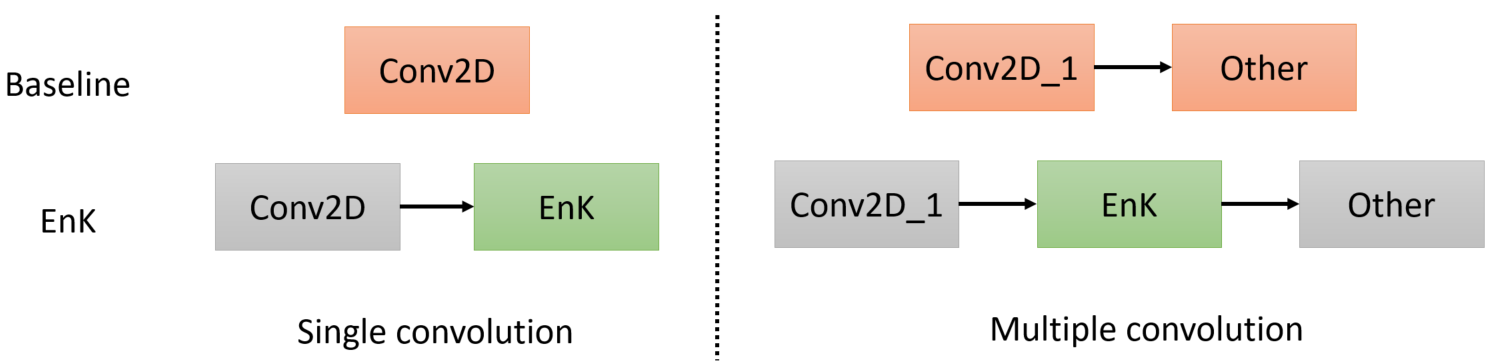}
\caption{Illustration of uses of the EnK with baseline models}
\label{Figure 3}
\end{figure}

We have used three baseline models. These baseline models have shown to generalize to different tasks in EEG signals with optimal performance in decoding and classification. EnK approach has been used with these baseline models by adding a layer after the first convolution layer (see Figure \ref{Figure 3}).
\begin{enumerate}
    \item EEGNet \cite{lawhern2018eegnet}  is a compact CNN architecture and contains an input block, two convolutional blocks, and a classification block. EEGNet replaces the traditional convolution operation with a depthwise separable convolution inspired by Xception \cite{chollet2017xception}.
    \item DeepConvNet \cite{schirrmeister2017deep} is deeper and hence has many more parameters. It consists of four convolutional blocks and a classification block. The first convolutional block is specially designed to handle EEG inputs, and the other three are standard convolution ones.
    \item ShallowConvNet \cite{schirrmeister2017deep} is a shallow version of DeepConvCNN, inspired by filter bank common spatial patterns \cite{ang2008filter}. Its first block is similar to the first convolutional block of DeepConvNet, but with a larger kernel, a different activation function, a different pooling approach, and a classification block.
\end{enumerate}

\subsection{Evaluation metrics and parameters settings}
The performance of EnK is compared with EEGNet, ShallowConvNet, and DeepConvNet. To compare, we have updated the same model with and without enabling the EnK layer and evaluated F1 score and accuracy. Figure \ref{Figure 3} demonstrates the architecture of models with EnK. For binary class, F1 score and accuracy are calculated as absolute values, but for multiclass, the F1 score is the weighted average for all labels. We also used Gaussian noise in place of EnK to compare the performance. \par

We have also compared the gradient weighted class visualization map (Grad-CAM) \cite{selvaraju2017grad} of EnK after the first convolution with the first layer output of baseline models for CC, pHRC, P300, and MRCP datasets. It is noted that we have used the output of DeepConvNet only for comparisons.

The proposed model is fitted using the Adam optimization algorithm with default parameters as defined in \cite{lawhern2018eegnet} \cite{schirrmeister2017deep}. We ran a maximum of 500 training epochs with batch sizes 16, 16, 8, and 4 for CC, pHRC, P300, and MRCP datasets. We trained the model with EEGNet, ShallowConvNet, and DeepConvNet with the original structure, with EnK  and Gaussian noise, enabled structure after the first convolution layer. Figure \ref{Figure 3} shows how EnK layer is used with the existing structure. We follow similar ways for the Gaussian noise layer. Each trained model has been saved together with weights, training loss, and training accuracy. All models were trained on a machine powered by NVIDIA Quadro P5000 GPU, with CUDA 9 and cuDNN v7, developed using Keras \footnote{Keras: https://keras.io/}.

\section{Results and discussion}

\subsection{Performance comparison with and without EnK}
As shown in Table \ref{Table 2}, we have compared the effect of EnK with standard EEGNet, ShallowConvNet, and DeepConvNet model. The results clearly showed that the EnK approach significantly improves the accuracy compared to the standard proposed model. The EnK outperforms in accuracy for CC dataset with state-or-art by 2.29\%, 1.55\%, ~1\% for EEGNet, ShallowConvNet, and DeepConvNet respectively. There was slight and comparable improvement found for pHRC and P300 dataset with state-of-art models. However, EnK shows a significant increase for MRCP dataset, with 11\% on average compare to EEGNet, ShallowConvNet, and DeepConvNet. These results suggest that EnK pushing CNN to learn additional time-dependency features wherever possible and vital.

\begin{table}[h]
\caption{The accuracy from EEGNet, ShallowConvNet, and DeepConvNet with/without EnK and with Gaussian noise after the first convolution layer  in CC, pHRC, P300, and MRCP datasets.}
\label{Table 2}
\begin{tabular}{lccccccccc}
\toprule
 & \multicolumn{3}{c}{EEGNet}  & \multicolumn{3}{c}{ShallowConvNet} & \multicolumn{3}{c}{DeepConvNet}  \\ 
\cmidrule{2-4} \cmidrule{5-7} \cmidrule{8-10} 
 Datasets       &  Org & EnK & Gauss
         &  Org & EnK & Gauss
	&  Org & EnK & Gauss                                                  \\    \hline
\textbf{CC}   & 0.7189                         & \cellcolor[HTML]{EFEFEF}\textbf{0.7418} & 0.7350         & 0.7263       & \cellcolor[HTML]{EFEFEF}\textbf{0.7418} & 0.7253                         & 0.7345                         & \cellcolor[HTML]{EFEFEF}\textbf{0.7418} & 0.7029                         \\
\textbf{pHRC} & 0.7856                         & \cellcolor[HTML]{EFEFEF}\textbf{0.7862} & 0.7856         & 0.7828       & \cellcolor[HTML]{EFEFEF}\textbf{0.7856} & 0.7842                         & \cellcolor[HTML]{EFEFEF}\textbf{0.7869} & \cellcolor[HTML]{FFFFFF}0.7856 & 0.7856                         \\
\textbf{P300} & \cellcolor[HTML]{EFEFEF}\textbf{0.9412} & \cellcolor[HTML]{EFEFEF}\textbf{0.9412} & 0.9216         & 0.8922       & \cellcolor[HTML]{EFEFEF}\textbf{0.9314} & \cellcolor[HTML]{EFEFEF}\textbf{0.9314} & 0.9216                         & \cellcolor[HTML]{EFEFEF}\textbf{0.9314} & \cellcolor[HTML]{EFEFEF}\textbf{0.9314} \\
\textbf{MRCP} & 0.4526                         & \cellcolor[HTML]{EFEFEF}\textbf{0.5625} & 0.5053         & 0.4842       & \cellcolor[HTML]{EFEFEF}\textbf{0.6000} & 0.4947                         & 0.4421                         & \cellcolor[HTML]{EFEFEF}\textbf{0.5579} & 0.5053                         \\
\bottomrule
\end{tabular}
\end{table}
As a better measure, we also looked at the F1 score to better understand how well EnK improving the performance compare to state-of-art models. The EnK improves the accuracy for CC dataset with state-or-art by 1.74\%, 1.22\%, compared to EEGNet, ShallowConvNet, and slight improvement compared to DeepConvNet. Similar to accuracy results, we also found a slight and comparable increase for pHRC and P300 dataset with EEGNet models. However, EnK shows again a significant improvement for MRCP dataset, with 12\% on average compare to EEGNet, ShallowConvNet, and DeepConvNet. These results again indicate the EnK is a better approach compared to state-of-art. 

\begin{table}[h]
\caption{The F1 score from EEGNet, ShallowConvNet, and DeepConvNet  with/without EnK and with Gaussian noise after the first convolution for CC, pHRC, P300, and MRCP datasets.}
\label{Table 3}
\begin{tabular}{lccccccccc}
\toprule
  & \multicolumn{3}{c}{EEGNet}  & \multicolumn{3}{c}{ShallowConvNet} & \multicolumn{3}{c}{DeepConvNet}  \\ 
\cmidrule{2-4} \cmidrule{5-7} \cmidrule{8-10} 
    Datasets   &  Org & EnK & Gauss
         &  Org & EnK & Gauss
	&  Org & EnK & Gauss                                                  \\    \hline
\textbf{CC}   & 0.8344                                  & \cellcolor[HTML]{EFEFEF}\textbf{0.8518} & 0.8465         & 0.8396       & \cellcolor[HTML]{EFEFEF}\textbf{0.8518} & 0.8398                         & 0.8468                         & \cellcolor[HTML]{EFEFEF}\textbf{0.8518} & 0.8218                                  \\
\textbf{pHRC} & 0.8798                                  & \cellcolor[HTML]{EFEFEF}\textbf{0.8801} & 0.8799         & 0.8782       & \cellcolor[HTML]{EFEFEF}\textbf{0.8799} & 0.8791                         & \cellcolor[HTML]{FFFFFF}0.8806 & \cellcolor[HTML]{EFEFEF}\textbf{0.8799} & \cellcolor[HTML]{EFEFEF}\textbf{0.8799} \\
\textbf{P300} & \cellcolor[HTML]{EFEFEF}\textbf{0.9423} & \cellcolor[HTML]{FFFFFF}\textbf{0.9412} & 0.9200         & 0.8991       & \cellcolor[HTML]{EFEFEF}\textbf{0.9333} & \cellcolor[HTML]{EFEFEF}0.9293 & 0.9200                         & \cellcolor[HTML]{EFEFEF}\textbf{0.9333} & \cellcolor[HTML]{FFFFFF}0.9320          \\
\textbf{MRCP} & 0.4501                                  & \cellcolor[HTML]{EFEFEF}\textbf{0.5621} & 0.4951         & 0.4816       & \cellcolor[HTML]{EFEFEF}\textbf{0.5804} & 0.4885                         & 0.4080                         & \cellcolor[HTML]{EFEFEF}\textbf{0.5564} & 0.3732                                  \\
\bottomrule
\end{tabular}
\end{table}

\subsection{Performance between EnK and Gaussian noise}
We have also evaluated if better performance is simply the result of noise inclusion within the dataset. To do so, Gaussian noise has been added after the first convolution layer (similar to EnK). Again, we have compared the results of Gaussian noise with EnK. We found that the EnK approach always outperforms the Gaussian noise approach, which proves that EnK is different from simply adding noise. However, for P300 and pHRC dataset with DeepConvNet model, EnK performance is comparable with Gaussian noise but significantly superior in terms of accuracy and F1 score for all four datasets. See Table \ref{Table 2} for detail results. \par

It is important to note that, EnK approach does not add any significant overhead to existing EEGNet, ShallowConvNet, and DeepConvNet. As shown in Table \ref{Table 4}, EnK approach always required an additional trainable parameter compare to state-of-art models. 

\begin{table}[h]
\caption{The number of trainable parameters with and without EnK approach with EEGNet, ShallowConvNet, and DeepConvNet for CC, pHRC, P300, and MRCP datasets}
\label{Table 4}
\centering
\begin{tabular}{lcccccc}
\toprule
  & \multicolumn{2}{c}{EEGNet}  & \multicolumn{2}{c}{ShallowConvNet} & \multicolumn{2}{c}{DeepConvNet}  \\ 
\cmidrule{2-3} \cmidrule{4-5} \cmidrule{6-7}
   Datasets  & Org                            & EnK                            & Org                            & EnK                            & Org                            & EnK                            \\
   \midrule
CC   & \cellcolor[HTML]{FFFFFF}180,794 & \cellcolor[HTML]{FFFFFF}180,795 & \cellcolor[HTML]{FFFFFF}112,642 & \cellcolor[HTML]{FFFFFF}112,643 & \cellcolor[HTML]{FFFFFF}199,527 & \cellcolor[HTML]{FFFFFF}199,528 \\
pHRC & \cellcolor[HTML]{FFFFFF}58,754  & \cellcolor[HTML]{FFFFFF}58,755  & \cellcolor[HTML]{FFFFFF}64,642  & \cellcolor[HTML]{FFFFFF}64,643  & \cellcolor[HTML]{FFFFFF}180,777 & \cellcolor[HTML]{FFFFFF}180,778 \\
P300 & \cellcolor[HTML]{FFFFFF}160,514 & \cellcolor[HTML]{FFFFFF}160,515 & \cellcolor[HTML]{FFFFFF}104,882 & \cellcolor[HTML]{FFFFFF}104,883 & \cellcolor[HTML]{FFFFFF}176,777 & \cellcolor[HTML]{FFFFFF}176,778 \\
MRCP & \cellcolor[HTML]{FFFFFF}44,244  & \cellcolor[HTML]{FFFFFF}44,245  & \cellcolor[HTML]{FFFFFF}55,604  & \cellcolor[HTML]{FFFFFF}55,605  & \cellcolor[HTML]{FFFFFF}171,479 & \cellcolor[HTML]{FFFFFF}171,480 \\
\bottomrule
\end{tabular}
\end{table}

\subsection{Grad-CAM comparison}
\begin{figure}[h]
\centering
\includegraphics[width=1\textwidth]{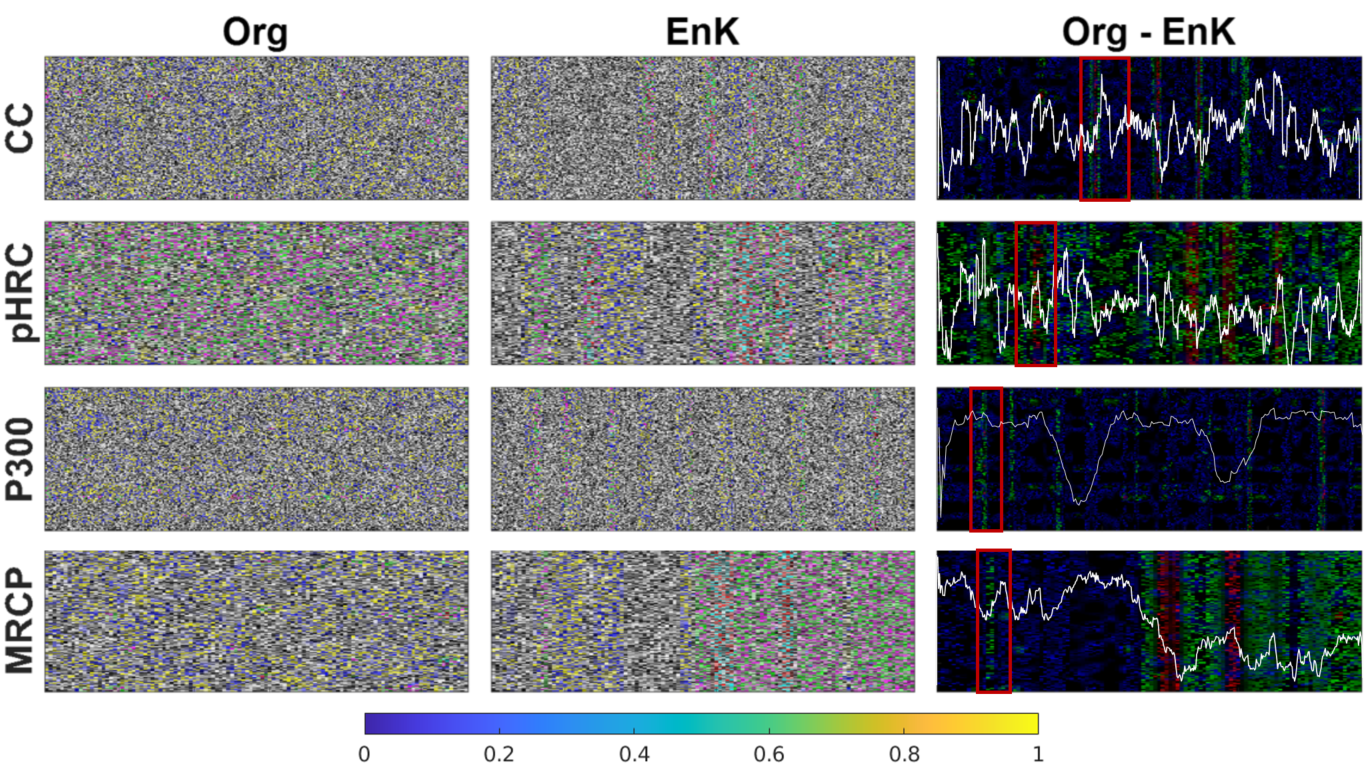}
\caption{Gradient-weighted Class Activation Mapping (Grad-CAM) from the original model (first column), with EnK (second column), and their difference (third column) for all datasets (CC, pHRC, P300, and MRCP). The last column is overlayed with original data (line graph) used to produce Grad-Cam results.}
\label{Figure 4}
\end{figure}

Our performance results show that the EnK approach outperforms the state-of-art, as well proven to be superior from simply adding noise. Such results imply that EnK is working as expected and learning time-dependency features. But to get a better answer, we have compared learned gradient between EnK approach with state-of-art models. Our results of Grad-CAM is shown in Figure \ref{Figure 4}. It can be seen from Figure \ref{Figure 4} (last column) that EnK is successfully able to introduce time-information in the data. This information can be seen as vertical lines representing the main features learned. According to \cite{singh2018visual} and \cite{aldini2019effect} for CC and pHRC data, the PEN and related features is majorly captured by EnK approach as shown in red box (see Figure \ref{Figure 4} for CC and pHRC). Similarly, P300 and MRCP can be seen to capture prominent features also shown in the red box in (see Figure \ref{Figure 4} for P300 and MRCP). \par

For simplicity, the raw data (line graph from a EEG signal) has been overlayed over the Figure \ref{Figure 4} (last column). In addition to traditional features, the EnK approach enabled further feature learning, which is one of the essential advantages of using raw data compared to hand-crafted features. The new features let model learn better about the problem and potentially reveal additional information. For example, in CC, features highlighted are indicating the N400 \cite{kutas2011thirty}. N400 is another vital feature in cognitive conflict kind of task. Similarly, MRCP is also showing more activity learned by the model. The model learn additional information from the remaining half part of the task in addition to the beginning. These learned features could be the source of new information unknown so far, particularly in cognitive neuroscience or related to other significant phenomena, ignores otherwise. These results provide further evidence that EnK has an advantage over the existing approach in learning time-dependency features. \par

Overall, EnK shows promising and significant improvement over the existing state-of-art models. EnK could be a potential tool for further investigation of new or existing phenomena in cognitive neuroscience and potentially beyond. This approach can be applied virtually in any existing and current CNN architectures to learn time-dependent features.

\section{Conclusion and future work}
In this work, we have introduced the EnK approach, which encodes time-information in CNNs. EnK introduces the time information in the data by modulating a sliding kernel over columns in convolution operation. EnK has been evaluated with various EEG signals from different paradigms with varying data sizes, channels, and sampling rates. EnK approach significantly outperforms the state-of-art models by leveraging the time-dependent features. In-addition, EnK showed a potential use case to explore new features and phenomena in EEG signals. Besides several advantages, EnK can be used with any existing model with negligible computational overhead and independent of the model's architecture. In future work, we plan to introduce non-linearity in EnK and further exploration with a wide variety of time-series signals.

\section{Broader Impact}

The EnK approach is independent of any CNN architecture and can be applied to and extend the functionality of any CNN models. It can be directly used with any CNN architecture proposed, and or in future with respect to time-series data. EnK is devised to improve the decoding and classification problem in EEG signals and theoretically also applicable to any physiological and non-physiological datasets. \par

The EnK approach is directly benefiting any stakeholder related to biomedical applications, and failure will not corrupt the critical systems. 

\small
\printbibliography
\end{document}